\documentclass[10pt,journal,compsoc]{IEEEtran}

\usepackage{booktabs} 
\usepackage{subcaption}
\usepackage{multirow}
\usepackage{enumitem}
\usepackage{array}
\usepackage{bm}
\usepackage{amsthm}
\usepackage{amsmath}
\usepackage{amsfonts}
\usepackage{threeparttable}

\usepackage{graphicx}
\usepackage{algorithm}
\usepackage{algpseudocode}
\usepackage{verbatim}
\usepackage{amssymb}
 \usepackage{mathtools}
\usepackage{xcolor}
\usepackage{amsthm}
\usepackage{url}

\usepackage[resetlabels,labeled]{multibib}
\newcites{R}{REFERENCES}
\newcites{A}{REFERENCES}

\newcommand{\LIANG}{}

\newcommand{\FinalOne}{}
\newcommand{\LCThree}{}

\begin{document}

\title{Region Embedding with Intra and Inter-View Contrastive Learning}

%
%

\author{Liang~Zhang,
        Cheng~Long,
        and~Gao~Cong
\IEEEcompsocitemizethanks{
\IEEEcompsocthanksitem L. Zhang, C. Long and G. Cong are with the Department of Computer Science, Nanyang Technological University, Singapore. E-mail: liang012@e.ntu.edu.sg, \{c.long, gaocong\}@ntu.edu.sg

Corresponding author: Cheng Long
}

 \thanks{This study is supported under the RIE2020 Industry Alignment Fund – Industry Collaboration Projects (IAF-ICP) Funding Initiative, as well as cash and in-kind contribution from Singapore Telecommunications Limited (Singtel), through Singtel Cognitive and Artificial Intelligence Lab for Enterprises (SCALE@NTU).
 This research is also supported by the Ministry of Education, Singapore, under its Academic Research Fund (Tier 2 Award MOE-T2EP20221-0013). 
 This research is also supported by the National Research Foundation, Singapore under its Industry Alignment Fund – Pre-positioning (IAF-PP) Funding Initiative. Any opinions, findings and conclusions or recommendations expressed in this material are those of the author(s) and do not reflect the views of the Ministry of Education, Singapore or National Research Foundation, Singapore.}
}

%
%

\markboth{Journal of \LaTeX\ Class Files,~Vol.~14, No.~8, August~2015}%
{Shell \MakeLowercase{\textit{et al.}}: Region Embedding with Intra and Inter-View Contrastive Learning}
%



\IEEEtitleabstractindextext{%
\begin{abstract}
Unsupervised region representation learning aims to extract dense and effective features from unlabeled urban data. While some efforts have been made for solving this problem based on multiple views, existing methods are still insufficient in extracting representations in a view and/or incorporating representations from different views. Motivated by the success of contrastive learning for representation learning, we propose to leverage it for multi-view region representation learning and design a model called ReMVC (Region Embedding with Multi-View Contrastive Learning) by following two guidelines: $i$) comparing a region with others within each view for effective representation extraction and $ii$) comparing a region with itself across different views for cross-view information sharing. We design the intra-view contrastive learning module which helps to learn distinguished region embeddings and the inter-view contrastive learning module which serves as a soft co-regularizer to constrain the embedding parameters and transfer knowledge across multi-views. We exploit the learned region embeddings in two downstream tasks named land usage clustering and region popularity prediction. Extensive experiments demonstrate that our model achieves impressive improvements compared with seven state-of-the-art baseline methods, and the margins are over 30\% in {\LCThree the} land usage clustering task. 
\end{abstract}

}

\maketitle

\IEEEdisplaynontitleabstractindextext

%
\IEEEpeerreviewmaketitle

\IEEEraisesectionheading{\section{Introduction}}
As basic elements of a city, regions are areas where people live, work, and entertain. Studying region representations can help us explore the functions and properties of cities, which is beneficial for many downstream applications such as land usage clustering \cite{zhang2020multi} and popularity prediction \cite{fu2019efficient}, and ultimately makes our cities smarter. 
In recent years, with the pervasiveness of mobile sensing technologies, a large amount of urban data such as Point-Of-Interests (POIs) and human mobility records become available. They provide great opportunities for addressing the region representation learning problem jointly from multi-views, where one view called the \emph{POI view} is based on POIs and the other one called the \emph{mobility view} is based on human mobility records. Many existing studies have been conducted on the region representation learning task based on these two views~\cite{fu2019efficient,zhang2019unifying,zhang2020multi,wang2017region,yao2018representing}. 
{\LCThree They} typically tackle the problem following two steps: intra-view feature extraction and multi-view embedding fusion. 

For intra-view feature extraction, two strategies have been studied. $i)$ One is to construct a region graph based on region correlations and learn representations {\LCThree with} graph embedding methods~\cite{zhang2020multi,wang2017region}. 
However, the graph is usually defined based on heuristic measures, which encode the region features manually instead of a learnable encoder. 
$ii)$ The other strategy addresses the mentioned shortcomings to some extent by learning a parameterized  encoder to reconstruct the region features~\cite{zhang2019unifying}. 
Nevertheless, the feature reconstruction strategy may 
focus on region details with little representative information. 
For example, considering a case where two regions both contain several residential buildings but serve as different functions reflected by some discriminate POIs such as bars and parks. With perfect POI reconstructions, these residential POIs {\LCThree would} dominate the model learning and 
then generate similar representations for these two regions (with different urban functions). 

%
%

For the multi-view embedding fusion, existing works exploit either concatenation~\cite{fu2019efficient,zhang2019unifying} or weighted average strategies~\cite{zhang2020multi}. 
However, the observations from different views should be highly correlated for the same region while less correlated for different regions, and the existing strategies can not preserve the cross-view correlations explicitly. 

%

Recently, the contrastive learning method has achieved great success in computer vision and natural language processing domains, demonstrating its superiority over other methods 
\cite{henaff2020data}. Motivated by its success, we propose to learn region representations by introducing comparison, which can help to capture abstract features and complex structures. 
Principally, a region may include some distinguished properties such as special POIs, which should be emphasised for intra-view feature extraction. We can capture this aspect easily by adopting the first guideline which is to \emph{compare a region with others within each view}. 

Furthermore, for a region, the multi-view data is generated due to the intrinsic region properties. For example, an entertainment region often includes certain POI types such as bars (in the POI view) and specific human mobility patterns such as large-scale flows in the evening (in the mobility view). 
Intuitively, only reasonable representations from multi-views can match each other. Therefore, we propose another guideline for multi-view embedding fusion which is to \emph{compare a region with itself across different views}.

By following the two guidelines, in this paper, we propose a general representation learning framework called ReMVC (Region Embedding with Multi-View Contrastive Learning) for region embedding based on two views, namely the POI view and the mobility view. 
For each view, we design the \emph{intra-view contrastive learning} module to extract region representations by maximizing the discrepancy between a region and other {\FinalOne dissimilar} regions. 
Furthermore, we design the \emph{inter-view contrastive learning} module to preserve the proposed cross-view correlations between different views, and the feature learning in one view can constrain the learning in the other view.
By using one view as the label for another, our model can also be regarded as transferring knowledge {\LCThree along} bi-directions.
Each view can learn how to represent the region by treating another view as the reference. In summary, we make the following contributions:

\begin{itemize}
{\LIANG{\item We design a multi-task learning framework called ReMVC which mainly consists of (1) an intra-view contrastive learning module for discriminate representations extraction; 
(2) and an inter-view contrastive learning module 
for inter-view cooperation and information sharing.
In fact, we are aware of no existing methods which use 
contrastive learning for both intra and inter-view
representation learning. 

\item We conduct extensive experiments 
and the results demonstrate that our model outperforms existing methods on two downstream tasks, and the margins are over 30\% in {\LCThree the} land usage clustering task.}} Our codes and data are available online \footnote{https://github.com/Liang-NTU/ReMVC}.

\end{itemize} 


\vspace{-0.1cm}
 \section{Related Work} \label{sec:rw}
\textbf{Region Representation Learning.}
Recently, a few studies propose to learn generic region representations from unlabeled urban data. To this end, the geo-tagged social media data and satellite images are exploited firstly. The study in \cite{hong2012discovering} exploits topic models to process twitter texts and the extracted topics can serve as region representation. Following studies extend the idea by using deep learning methods. For instance, TNE \cite{paul2021semantic} {\LCThree defines} a graph learning framework for geo-tagged social media data. Tile2Vec \cite{jean2019tile2vec} regards a region as satellite images and learns region embeddings with image representation {\LCThree learning} methods. M2G \cite{huangm3g} proposes a multi-task learning framework based on both geo-tagged check-ins and satellite images. 




Other studies propose approaches that merely use easily available data including human mobility patterns and POI attributes to learn representations of regions.
Along this line, \cite{wang2017region} and \cite{wang2018learning} construct transition graphs between regions based on human mobility patterns and use path sampling and AutoEncoder to extract region representations. CGAL \cite{zhang2019unifying} considers both mobility connectivity and geographic distance to construct POI-POI graphs for each region and develops an ``encode-decode'' based framework. \cite{zhang2020multi} further extends CGAL by replacing the AutoEncoder with 
cross-view GNNs. However, these approaches follow the similar learning strategy by using {\LCThree a} \emph{reconstruction} framework and a simple embedding fusion method, which may lead to ineffective region representations, as explained in the introduction section. Different from these studies, we introduce the contrastive learning for both intra and inter-view to enhance discriminate embedding learning and multi-view cooperation. 

\noindent\textbf{Contrastive Learning.}
 Contrastive learning is an emerging machine learning paradigm which aims to learn 
 data representations form unlabeled raw data \cite{gidaris2018unsupervised}.
The typical contrastive models learn data representation through comparing similar sample pairs with dissimilar ones {\LCThree and adopting} the Noise Contrastive Estimation (NCE) objective. {\FinalOne{It is commonly used for visual and video feature extraction \cite{chen2020simple}.}} 
For example, simply asking the network to recognize the image orientation and invariance to data augmentation \cite{gidaris2018unsupervised} can {\LCThree learn high-quality} image representations which can {\LCThree be used in} downstream tasks such as image {\LCThree classification}. Nowadays, contrastive learning has also been applied to other domains such as graph embedding and point {\LCThree set} representations \cite{arsomngern2021towards}. For instance, based on the mutual information between each node and the local graph structure, pre-trained GNNs \cite{velivckovic2018deep} can learn robust node and graph representations. 
In this paper, we propose a 
contrastive learning framework for region presentation learning based on POI data and mobility data, which is the first of it kind.



\noindent\textbf{Multi-view Representation Learning.}
{\FinalOne{There is a rich literature of multi-view representation learning: most of the methods were designed for specific views of data 
such as images \cite{yang2019skeletonnet}, graphs \cite{zhou2022multiview} and heterogeneous graphs \cite{shao2021heterogeneous}}}, yet few is generally applicable across different views of data. These methods can be roughly categorized as \emph{representation alignment methods} and \emph{representation fusion methods} \cite{li2018survey}: the former learn multi-view representations jointly with the alignment among different views used as regularizers/constraints and the latter learn representations of individual views separately and then fuse the representations as those for the multi-views. 


Our method is a representation alignment method and differs from existing methods in: (1) it uses 
contrastive learning for both intra-view and inter-view representations learning for the first time; and (2) it is tailored for region data with POI view and mobility view.

\vspace{-0.3cm}
\section{Problem Formulation}
\label{sec:pro}
Following existing studies~\cite{zhang2020multi}, 
we partition the space into a set of $L$ non-overlapping regions, denoted by $\mathcal{R} = \{r_1, r_2, ..., r_L\}$. 
In this paper, we consider two views for region representation learning, namely the POI view which is based on the POIs in a region and the mobility view which is based on human mobility records.



\noindent\textbf{POI view.} The region attributes are inherent geographic features of each region. A certain type of attribute can be represented as $\mathbf{f}_k$ for the $k$-th region, where $\mathbf{f}_k \in \mathbb{R}^{F}$ and $F$ is the number of feature dimensions. In this paper, 
we use POI categories as region attributes by following \cite{fu2019efficient,zhang2020multi}. Each dimension in $\mathbf{f}_k$ denotes the ratio of POIs with the corresponding category in the $k$-th region. 

\noindent\textbf{Mobility view.} We define the mobility flow data as a set  $\mathcal{T} = \{t_1, t_2, ..., t_n\}$ of trips. Each trip $t_i$ is a tuple consisting of {\LCThree the} source and destination {\LCThree locations} of the trip as well as the time information, which is denoted as $(r_s, r_d, h)$. We map the source and destination to {\LCThree their} corresponding regions. Then we can define two heat maps $\mathbf{MS}_k$ and $\mathbf{MD}_k \in \mathbb{R}^{H \times L}$ for the $k$-th region, where $H$ is the number of time slices and $L$ is the number of regions. {\LIANG{The element  $MS^{i,j}_k$ for a region $r_k$ is defined as $MS^{i,j}_k = |\{(r_s, r_d, h) | h=i, r_s = r_j, r_d = r_k\}|$. Similarly, by 
specifying $r_k$ as the source and other regions $r_j$ as the destination, we can obtain another matrix $\mathbf{MD}_k$ for the $k$-th region.}}

\noindent\textbf{Region representation learning problem.} Given heat maps $\mathbf{MS}$, $\mathbf{MD}$ and region attributes $\mathbf{f}$ for regions in $\mathcal{R}$, the problem is to learn 
a latent embedding {\LCThree of each region}, which is denoted as $\mathbf{E} = \{\mathbf{e}_1, \mathbf{e}_2, ..., \mathbf{e}_L | \mathbf{e}_k \in \mathbb{R}^{d} \}$ with $d$ as the embedding size.

\vspace{-0.2cm}
\section{Methodology}
\begin{figure}[t!]
	\centering
	\includegraphics[width=0.41\textwidth]{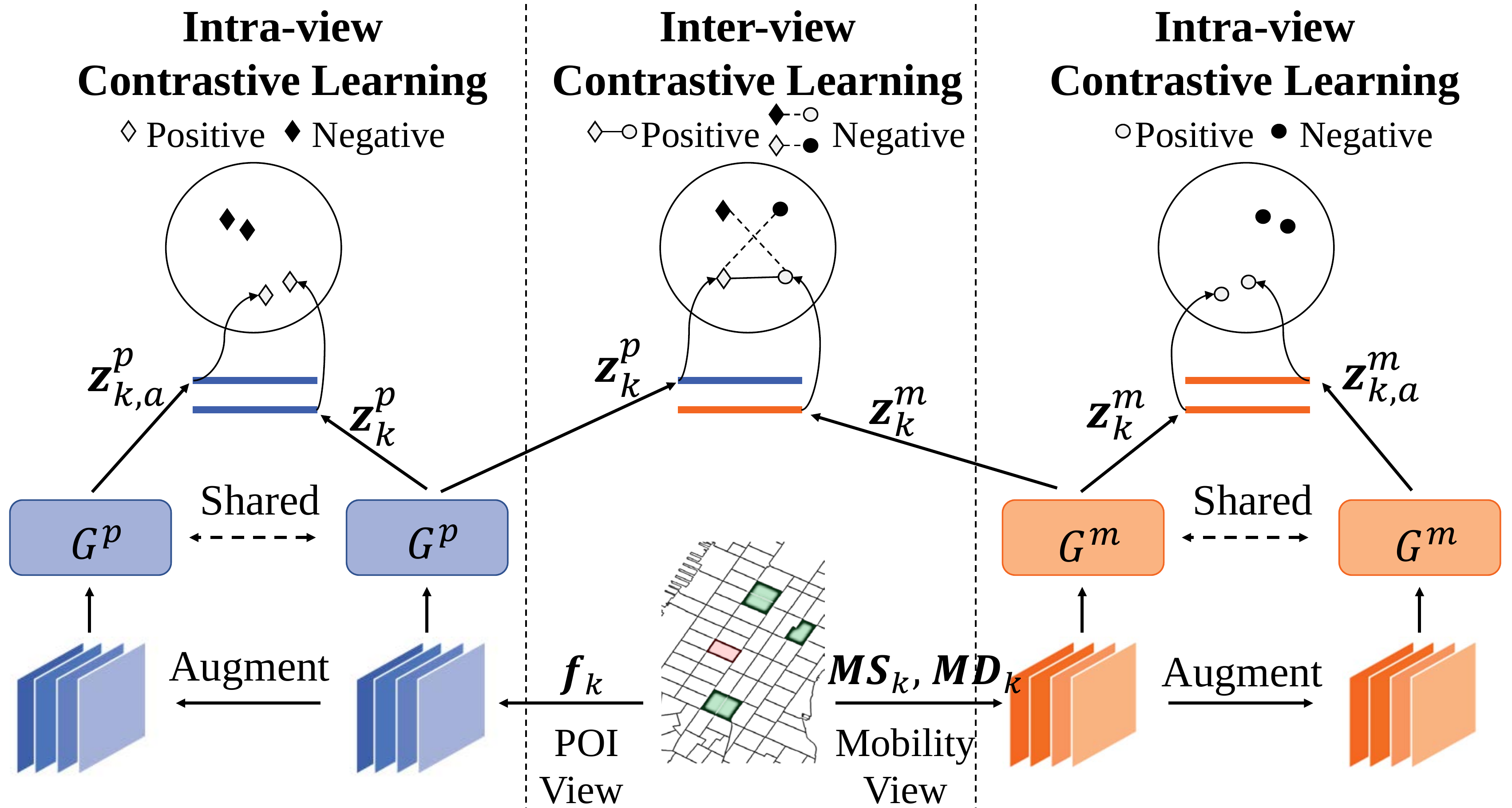}
	\caption{The architecture of ReMVC.}
	\label{fig:framework}
	\vspace{-0.6cm}
\end{figure}

In this section, we propose a novel deep neural network called ReMVC (Region Embedding with Multi-View Contrastive Learning) to cope with the region representation learning problem. The overview of the model architecture is shown in Figure \ref{fig:framework}, which has two important modules: the \emph{intra-view contrastive learning} 
to discriminate representations extracted from both POI and human mobility views, and the \emph{inter-view contrastive learning} 
to deal with effective multi-view embedding fusion. Our key innovation lies in incorporating contrastive learning in both views by following two guidelines. 
The first guideline is to compare a region with other regions within each view, which is followed by the intra-view contrastive learning module.
The second guideline is to compare a region with itself across different views, which is followed by the inter-view contrastive learning module.






\vspace{-0.3cm}
\subsection{Intra-view Contrastive Learning}


{\LCThree Consider the POI view first.}
The POI features to be encoded are defined as $\mathbf{f}_k$ for the $k$-th region $r_k$. 
We define an augmentation function $\varphi(\mathbf{f}_k,a)$, where $a$ is sampled from a set of data augmentation strategies $\mathcal{A}_p$. {\FinalOne{We design three POI level augmentations for $\mathcal{A}_p$: (1) Random Insertion: insert for each POI in region $r_k$ a random POI with probability $p$; (2) Random Deletion: drop each POI with probability $p$; and (3) Random Replacement: replace each POI with a random one with probability $p$.}}

Then the positive region set $\mathcal{P}_k$ {\LCThree of $r_k$ is} defined as $\mathcal{P}_k = \{ \mathbf{f}_{k,a} | a \in \mathcal{A}_p \}$ where the positive samples $\mathbf{f}_{k,a}$ are region augmentations $\varphi(\mathbf{f}_k,a)$, which is based on the principle that a region is similar to itself with some augmentations. 
The negative set $\mathcal{N}_k$ is defined as $\mathcal{N}_k = \{\mathbf{f}_n | n \neq k \}$  with $|\mathcal{N}_k|=N_p$, where $N_p$ is the predefined negative size for {\LCThree the} POI view and negative samples $\mathbf{f}_n$ are randomly sampled based on the POI feature distance distribution. For a candidate region $r_n$, its sampling probability is defined as the normalization of its POI feature distance from the anchor region $r_k$. The greater the POI feature distance, the more likely the region $r_n$ is sampled as a negative sample for {\LCThree the} POI view. 



For the $k$-th region, we then design a POI encoder $G^p$ to transform the coarse-grained POI features $\mathbf{f}_k$ to a latent embedding $\mathbf{z}^p_k$.  In this paper, we exploit the most commonly used structure Multi-Layer Perceptions (MLPs) to implement $G^p$ following the region representation studies in \cite{fu2019efficient}, \cite{wang2018learning} and \cite{zhang2019unifying}. Note that 
{\LCThree other}
feature extractors can {\LCThree possibly} be used in our framework.
Then, we have,
\begin{equation}
\label{eq:gp}
\mathbf{z}^p_k = G^p(\mathbf{f}_k) = \text{MLP}(\mathbf{f}_k).
\end{equation}

Given the POI encoder $G^p$, we obtain $\mathbf{z}^p_k = G^p(\mathbf{f}_k)$ and $\mathbf{z}^p_{k,a} = G^p(\mathbf{f}_{k,a})$ where $\mathbf{z}^p_k$ and $\mathbf{z}^p_{k,a}$ denote the learned representations for the $k$-th region $r_k$ in {\LCThree the} POI view and its feature augmentation, respectively. We then follow the first guideline and contrast the two groups of embeddings by adopting the famous InfoNCE loss \cite{han2020self}. 

{\FinalOne{
\vspace{-0.4cm}
\begin{equation}
\label{eq:poi}
\begin{aligned}
 \mathcal{L}&_{poi}  = \sum_ {r_k \in \mathcal{R}} \left[ - \log \sum_{a \in \mathcal{A}_p} D^{ia}(\mathbf{z}^p_k, \mathbf{z}^p_{k,a}) \right. + \\
  & \left. \log \left( \sum_{a \in \mathcal{A}_p} D^{ia}(\mathbf{z}^p_k, \mathbf{z}^p_{k,a}) + \sum_{n \in \mathcal{N}_k} D^{ia}(\mathbf{z}^p_k, \mathbf{z}^p_n) \right) \right].
\end{aligned}
\end{equation}
\vspace{-0.2cm}
}}

Here, the $\mathbf{z}^p_{k,a}$ is the positive embedding and $\mathbf{z}^p_{n}$ denotes the negative embedding. The function $D^{ia}$ is an intra-view discriminator that takes two vectors as {\LCThree inputs} and then scores the similarity between them. We simply implement it as the inner product as follows, where $\tau$ refers to the temperature parameter.

{\FinalOne{
\vspace{-0.25cm}
\begin{equation}
\begin{aligned}
D^{ia}(\mathbf{z}^p_k, \mathbf{z}^p_{k,a}) = \exp \left(\frac{\mathbf{z}^p_k \cdot \mathbf{z}^p_{k,a}}{\tau} \right).
\end{aligned}
\end{equation}
\vspace{-0.25cm}
}}

{\FinalOne{Similarly, we define the augmentation function $\varphi(\mathbf{MS}_k, \mathbf{MD}_k,a)$ and transformation set $\mathcal{A}_m$ for the mobility view, where $\mathcal{A}_m$ is implemented
based on an operation called Noise Injection: add to each element in heatmaps $\mathbf{MS}_k$ and $\mathbf{MD}_k$ a Gaussian noise with mean $0$ and variance $\sigma$.}}
%
%
%
We then 
define a positive set $\mathcal{P}_k = \{(\mathbf{MS}_{k,a},\mathbf{MD}_{k,a}) | a \in \mathcal{A}_m \}$ which is obtained based on region augmentations and a negative set $\mathcal{N}_k = \{(\mathbf{MS}_n,\mathbf{MD}_n) | n \neq k\}$ with $|\mathcal{N}_k|=N_m$, which is randomly sampled.
Similarly, we define the sampling probability {\LCThree of each candidate region} based on {\LCThree its} mobility feature distance from the anchor region $r_k$. 
$N_m$ is the predefined negative size for mobility view. The mobility view encoder $G^m$ is also implemented {\LCThree as an} MLP network as follows. 

\begin{equation}
\label{eq:gm}
\begin{aligned}
    \mathbf{z}^m_k & = G^m(\mathbf{MS}_k, \mathbf{MD}_k) \\ 
& = \text{Avg}(\text{MLP}(\mathbf{MS}_k), \text{MLP}(\mathbf{MD}_k)).
\end{aligned}
\end{equation}


Afterwards, based on the learned positive embedding $\mathbf{z}^m_{k,a} = G^m(\mathbf{MS}_{k,a}, \mathbf{MD}_{k,a})$ and negative embedding $\mathbf{z}^m_n = G^m(\mathbf{MS}_n, \mathbf{MD}_n)$, the intra-view contrastive signals for mobility view can be formulated as follows. 

{\FinalOne{
\vspace{-0.25cm}
\begin{equation}
\label{eq:mob}
\begin{aligned}
 \mathcal{L}&_{mob} =  \sum_ {r_k \in \mathcal{R}} \left[ - \log \sum_{a \in \mathcal{A}_m} D^{ia}(\mathbf{z}^m_k, \mathbf{z}^m_{k,a}) \right. + \\
 & \left. \log \left( \sum\limits_{a \in \mathcal{A}_m} D^{ia}(\mathbf{z}^m_k, \mathbf{z}^m_{k,a}) + \sum_{n \in \mathcal{N}_k} D^{ia}(\mathbf{z}^m_k, \mathbf{z}^m_n) \right) \right].
\end{aligned}
\end{equation}
\vspace{-0.9cm}
}}

\subsection{Inter-view Contrastive Learning}
We learn two groups of view-specific region embeddings separately via an intra-view contrastive learning module. To enhance the cooperation between different views and capture {\LCThree a} region’s intrinsic properties, we follow the second guideline, which is to compare a region with itself based on different views.
With this component, the feature learning in one view can constrain the learning in the other view, and hence it serves as a soft co-regularizer for the entire model. Conceptually, only good representations in both views can match each other.

Recall that, for the $k$-th region $r_k$, we obtain its region representations in each view as $\mathbf{z}^p_k$ and $\mathbf{z}^m_k$. Intuitively, the two embeddings can be the ground-truth of each other for self-supervised learning, and this one-to-one mapping can be regarded as the label augmentation. If two region representations {\LCThree correspond to those of} the same region, we label this pair as a positive pair, otherwise a negative pair. Specifically, for the $k$-th region, the positive pair should be the representations for the same region within different views formulated as ($\mathbf{z}^p_k$, $\mathbf{z}^m_{k}$). Meanwhile, by fixing the embedding of $r_k$ in one view and sampling other region representations from another view, we can obtain two groups of representations as $\mathcal{Z}^p = \{ \mathbf{z}^p_n | n \neq k\}$  with $|\mathcal{Z}^p|=N_i $ and $\mathcal{Z}^m = \{ \mathbf{z}^m_n | n \neq k\}$ with $|\mathcal{Z}^m|=N_i$, where $N_i$ is the predefined negative size for inter-view. 
Then the negative pairs are extracted as ($\mathbf{z}^p_k$, $\mathbf{z}^m_{n}$) and ($\mathbf{z}^p_n$, $\mathbf{z}^m_{k}$). The InfoNCE loss function for inter-view is defined as follows.

{\FinalOne{
\vspace{-0.25cm}
\begin{equation}
\label{eq:inter}
\begin{aligned}
 \mathcal{L}&_{inter} = \sum_ {r_k \in \mathcal{R}} \left[ - \log D^{ir}(\mathbf{z}^p_k, \mathbf{z}^m_{k}) \right. + \\
  & \left. \log \left( \splitfrac{ D^{ir}(\mathbf{z}^p_k, \mathbf{z}^m_{k}) + \sum_{\mathbf{z}^m_n \in \mathcal{Z}^m} D^{ir}(\mathbf{z}^p_k, \mathbf{z}^m_{n})} {+ \sum_{\mathbf{z}^p_n \in \mathcal{Z}^p} D^{ir}(\mathbf{z}^p_n, \mathbf{z}^m_k)} \right) \right].
\end{aligned}
\end{equation}
\vspace{-0.25cm}
}}

%
%
%
%
%
Here, $D^{ir}$ is an inter-view discriminator that evaluates the matching scores between two input region embeddings, which is different from $D^{ia}$ in {\LCThree the} intra-view contrastive learning for similarity measurement. 
Specifically, we implement it with a fully connected neural network by formulating the cross-view matching as a classification task. 

{\FinalOne{
\vspace{-0.35cm}
\begin{equation}
\begin{aligned}
D^{ir}(\mathbf{z}^p_k, \mathbf{z}^m_{k}) = \exp(\text{ReLU}(\mathbf{W} (\mathbf{z}^p_k \mathbin\Vert \mathbf{z}^m_{k}) + \mathbf{b})).
\end{aligned}
\end{equation}
\vspace{-0.25cm}

Here, $\mathbf{z}^p_k$ {\LCThree and} $\mathbf{z}^m_{k}$ are POI view and mobility view representations for region $r_k$, {\FinalOne respectively}, $\mathbin\Vert$ refers to the concat operation, and $\mathbf{W}$ and $\mathbf{b}$ are neural network parameters.}} Our model can also be regarded as transferring knowledge {\LCThree along} bi-directions by using one view as the label for {\LCThree the other}, which helps the intra-view {\LCThree module} obtain information from each other to improve feature extractions. 

\vspace{-0.2cm}
\subsection{Multi-task Learning}
Finally, we unify the different pre-tasks into a multi-task learning framework. 
The joint learning objective function is as follows, where $\alpha$ and $\beta$ are parameters controlling the balance between intra and inter-view contrastive learning.  

\begin{equation}
\label{eq:loss}
\begin{aligned}
 \mathcal{L} =  \mathcal{L}_{mob} + \alpha \cdot \mathcal{L}_{poi} + \beta \cdot \mathcal{L}_{inter}.
\end{aligned}
\end{equation}

After the model training, when given a new region $r_k$ with mobility heat maps $\mathbf{MS}_k$, $\mathbf{MD}_k$ and POI attributes $\mathbf{f}_k$, we can obtain the final region representation as follows. 

\vspace{-0.2cm}
\begin{equation}
\begin{aligned}
 \mathbf{e}_k =  G^p(\mathbf{f}_k) \mathbin\Vert G^m(\mathbf{MS}_k, \mathbf{MD}_k).
\end{aligned}
\end{equation}
\vspace{-0.6cm}

\vspace{-0.2cm}

\newcommand{\tabincell}[2]{\begin{tabular}{@{}#1@{}}#2\end{tabular}}  
\begin{table}[t!]
	\renewcommand\arraystretch{0.99}
	\centering
	\caption{Data Description.}
	\scriptsize
	
	\resizebox{0.67\linewidth}{!}{
		\begin{threeparttable}
			\begin{tabular}{c|c}
				\toprule
				Dataset & Description  \\
				\midrule
				Census blocks & \tabincell{c}{Boundaries of 270 regions split \\ by streets in Manhattan, New York.}  \\  
				\midrule
				Taxi trips  & \tabincell{c}{Around 10 million taxi trip records \\ during one month in the studied area.}  \\
				\midrule
			    POI data & \tabincell{c}{Over 17,000 POIs across over \\ 200 fine-grained categories} \\ \bottomrule
			\end{tabular}
		\end{threeparttable}
	}
	\label{tab:data}
	\vspace{-0.3cm}
\end{table}

\vspace{-0.1cm}
\section{Experiment}
We empirically evaluate our model with two important downstream applications: land usage clustering and region popularity prediction. The land usage clustering task is to cluster regions into different groups such as central business district (CBD) and the region popularity prediction task is to infer the check-in count in each region. Both of them are commonly used in existing studies of region embedding, e.g., land usage clustering is used in~\cite{zhang2020multi,yao2018representing} and region popularity prediction is used in~\cite{fu2019efficient,zhang2019unifying}.

\vspace{-0.2cm}
\subsection{Experiment Settings}
\textbf{Dataset.} We collect several real-world datasets of New York City from NYC open data website \footnote{https://opendata.cityofnewyork.us/}. The statistics of each dataset are reported in Table~\ref{tab:data}.
%
The first dataset is the boundary data. We use the city boundaries designed by the US Census Bureau \footnote{https://www.census.gov/data.html} for region embedding learning. Following the experimental settings in ~\cite{zhang2020multi}, we use the Manhattan borough as our studied area.
%
The second dataset is the human mobility data. We use complete taxi trip records of August 2013 as our training data. Each trip record contains the pick-up and drop-off locations and time stamps. For each time stamp, we map it to an hourly-long bin in one day to construct the mobility heat maps. 
%
%
The last dataset is the NYC POI data provided by \cite{yang2014modeling}, which includes over 17,000 POIs across over 200 fine-grained categories. We regard these locations' categories as POI attributes for each region following the work in~\cite{zhang2020multi}.





\noindent \textbf{Evaluation Metrics.} We use three metrics, namely Normalized Mutual Information (NMI), Adjusted Rand Index (ARI) and F-measure, for the land usage clustering task following \cite{yao2018representing}. And we use three metrics, namely Mean Absolute Error (MAE), Root Mean Square (RMSE), and the Coefficient of Determination ($R^2$), for the region popularity prediction task following \cite{zhang2020multi}. 
The detailed definitions of these metrics are presented {\LCThree in a technical report \footnote{\label{ft:fool_note}\url{https://github.com/Liang-NTU/Techinical_Report_ReMVC/blob/main/22-TKDE-ReMVC-report.pdf}}}. 

\noindent \textbf{Baseline Methods.}  We compare our method with the following 7 baselines, which are existing methods of region embedding based on mobility patterns and/or POI features. \textbf{POI} merely uses the POI categories of a region with {\LCThree the} TF-IDF method as the region embedding. \textbf{HDGE}  \cite{wang2017region} learns region representations by path sampling on a multi-layer mobility graph and spatial graph. \textbf{ZE-Mob} \cite{yao2018representing} learns region representations by employing mutual information matrix decomposition, which considers the region co-currency relations. \textbf{MV-PN} 
\cite{fu2019efficient} learns region embedding by compressing multi-view POI networks. \textbf{CGAL} \cite{zhang2019unifying} extends MV-PN by {\LCThree employing an} adversarial training method. \textbf{MVURE} \cite{zhang2020multi} learns region representations by performing graph convolution on multiple homogeneous graphs. 
\textbf{M2GRL} \cite{wang2020m2grl} is a general multi-view embedding method which learns view-specific representations by graph reconstruction. Among these baselines, POI, HDGE and ZE-Mob are all {\LCThree single-view based} methods which merely use the POI categories or human mobility patterns. MV-PN, CGAL, MVURE and M2GRL are multi-view based methods using both POI features and mobility patterns.
The implementation details are presented {\LCThree in our technical report \footnotemark[4].}


\begin{table}[t!]
	\renewcommand{\arraystretch}{1.1} 
	\centering 
	\scriptsize
	\caption{Performance on Two Downstream Tasks.}
	\resizebox{0.84\linewidth}{!}{
		\begin{tabular}{c|ccc|ccc}
			\toprule
			\multirow{2}{*}{Method} & \multicolumn{3}{c|}{{\it Land Usage Clustering}} & \multicolumn{3}{c}{{\it Popularity Prediction}}  \\
			\cmidrule{2-7}
			& NMI & ARI & F-measure & MAE & RMSE & $R^2$  \\ \midrule
			POI (TF-IDF) & 0.389 & 0.039 & 0.069 & 281.37 & 417.10 & 0.288 \\
			HDGE \cite{wang2017region} & 0.587 & 0.117 & 0.121 & 216.98 & 339.28 & 0.529   \\
			ZE-Mob \cite{yao2018representing} & 0.466 & 0.076 & 0.108 & 238.34 & 341.71 & 0.522 \\
			\midrule
		    MV-PN \cite{fu2019efficient} & 0.407 & 0.057 & 0.085 & 276.29 & 419.68 & 0.279 \\
		    CGAL \cite{zhang2019unifying} & 0.411 & 0.078 & 0.102 & 275.03 & 415.79 & 0.293 \\
		    MVURE \cite{zhang2020multi} & 0.691 & 0.337 & 0.351 & 222.55 & 336.39 & 0.537 \\
		    M2GRL \cite{wang2020m2grl} & \underline{0.696} & \underline{0.361} & \underline{0.383} & \underline{194.01} & \underline{303.89} & \underline{0.625}  \\
			\cmidrule{1-7}
			ReMVC & \textbf{0.762} & \textbf{0.474} & \textbf{0.488} & \textbf{189.92} & \textbf{296.26} & \textbf{0.643} \\ 
			\bottomrule
	\end{tabular}}
	\label{tab:CR}
	\vspace{-0.2cm}
\end{table}

\vspace{-0.2cm}
\subsection{Land Usage Clustering}

\begin{figure}[t!]
	\centering
	\setlength{\fboxrule}{0.4pt}
    \setlength{\fboxsep}{0.12cm}
	\begin{subfigure}[b]{0.093\textwidth}
		\fbox{\includegraphics[width=\textwidth]{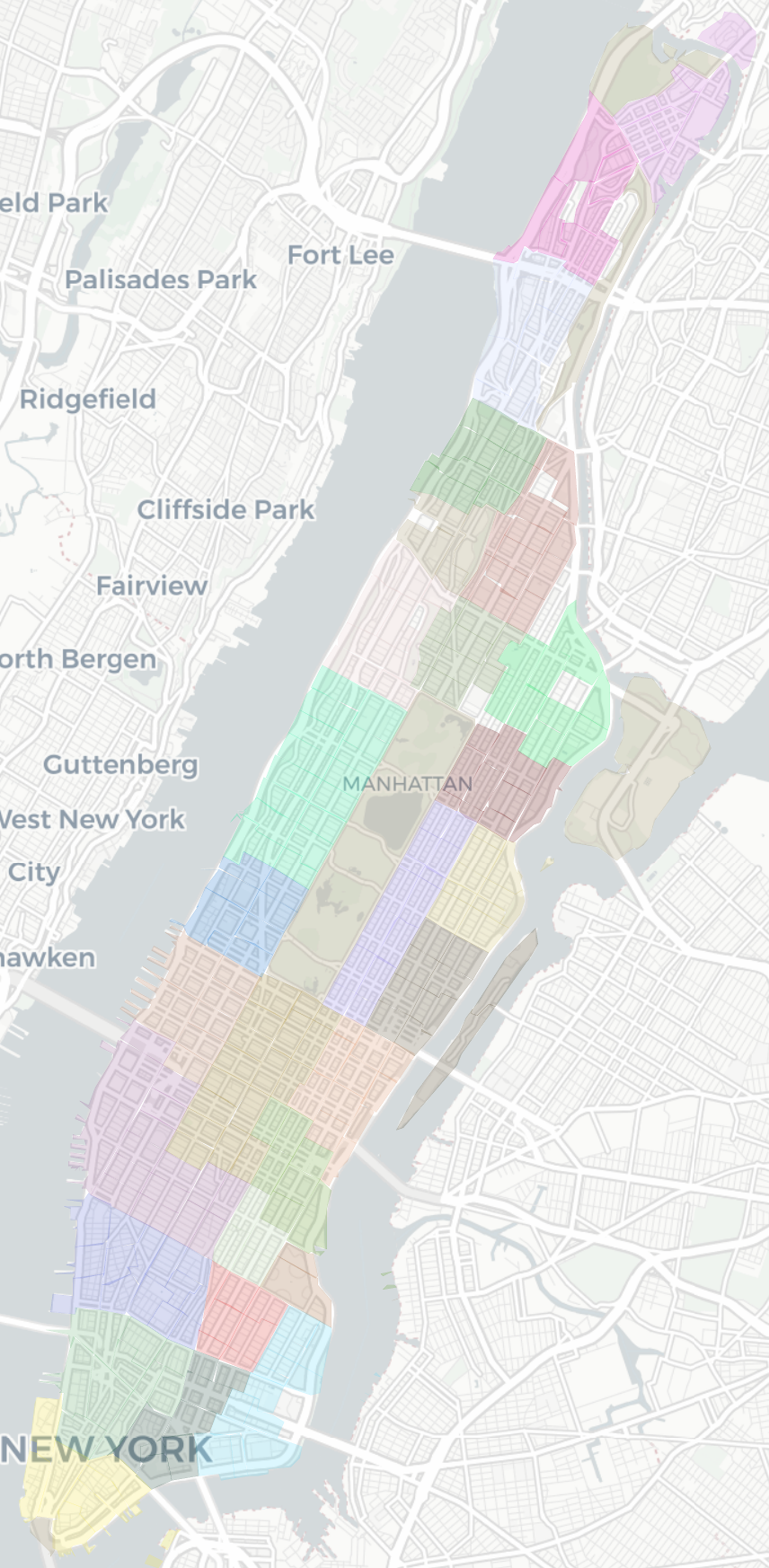}}
		\caption{\fontsize{6.6pt}{6.6pt}\selectfont{Labels}}
		\label{fig:main_labels}
	\end{subfigure}
	\hspace{0.01\textwidth}
	\begin{subfigure}[b]{0.093\textwidth}
		\fbox{\includegraphics[width=\textwidth]{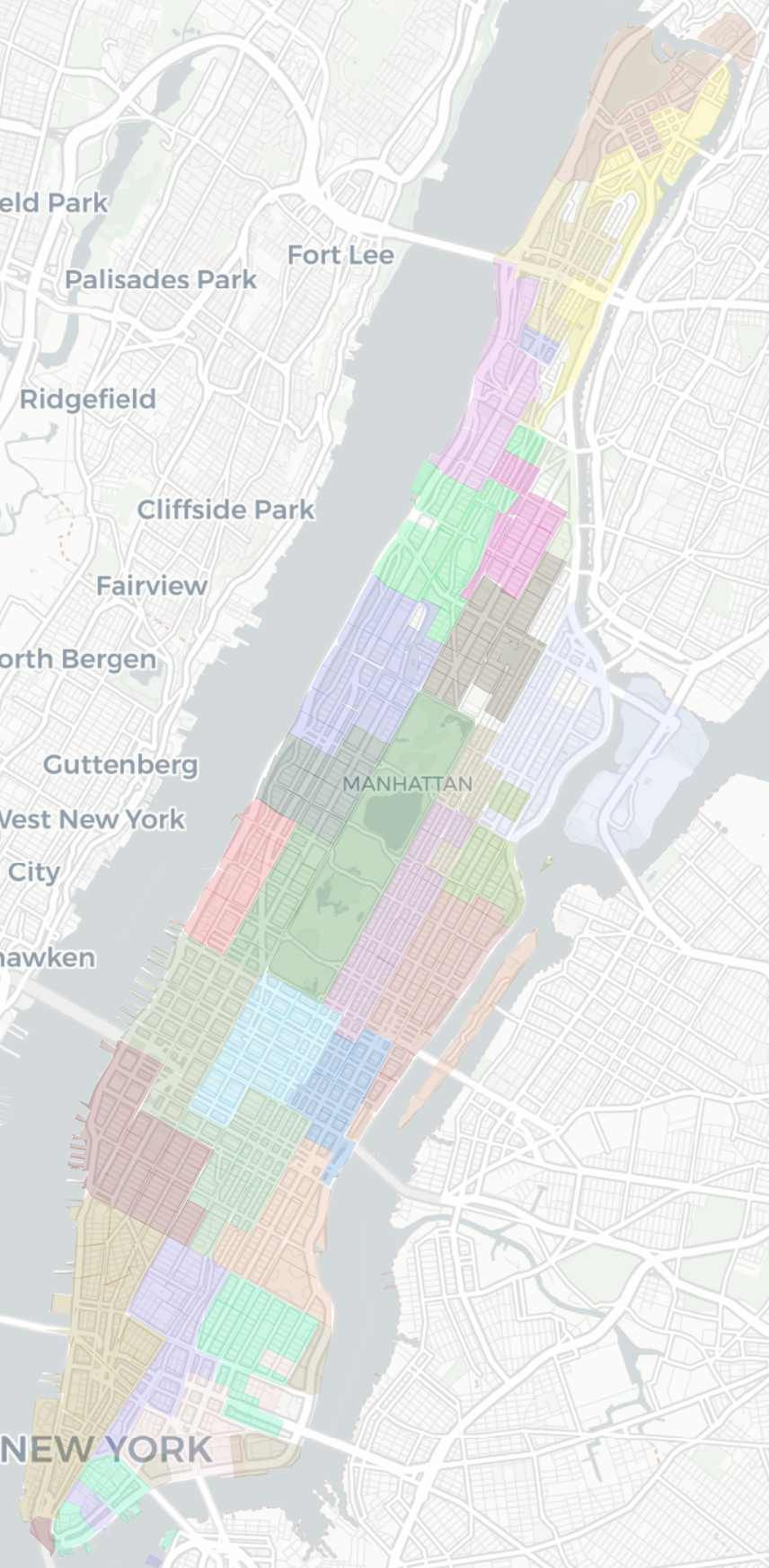}}
		\caption{\fontsize{6.6pt}{6.6pt}\selectfont{ReMVC}}
		\label{fig:ours}
	\end{subfigure}
	\hspace{0.01\textwidth}
	\begin{subfigure}[b]{0.093\textwidth}
		\fbox{\includegraphics[width=\textwidth]{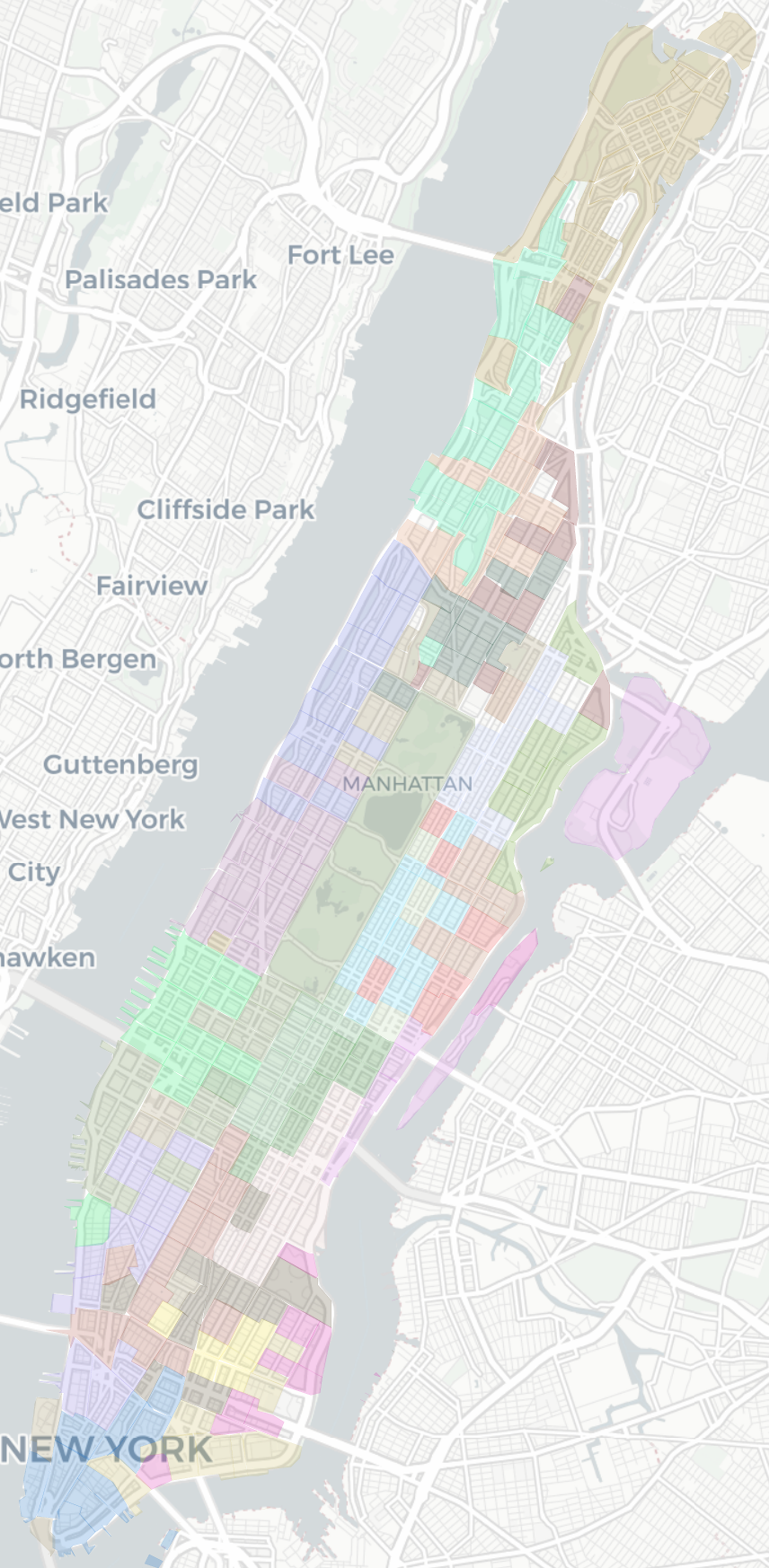}}
		\caption{\fontsize{6.6pt}{6.6pt}\selectfont{M2GRL \cite{wang2020m2grl}}}
		\label{fig:M2GRL}
	\end{subfigure}
	\hspace{0.01\textwidth}
	\begin{subfigure}[b]{0.093\textwidth}
		\fbox{\includegraphics[width=\textwidth]{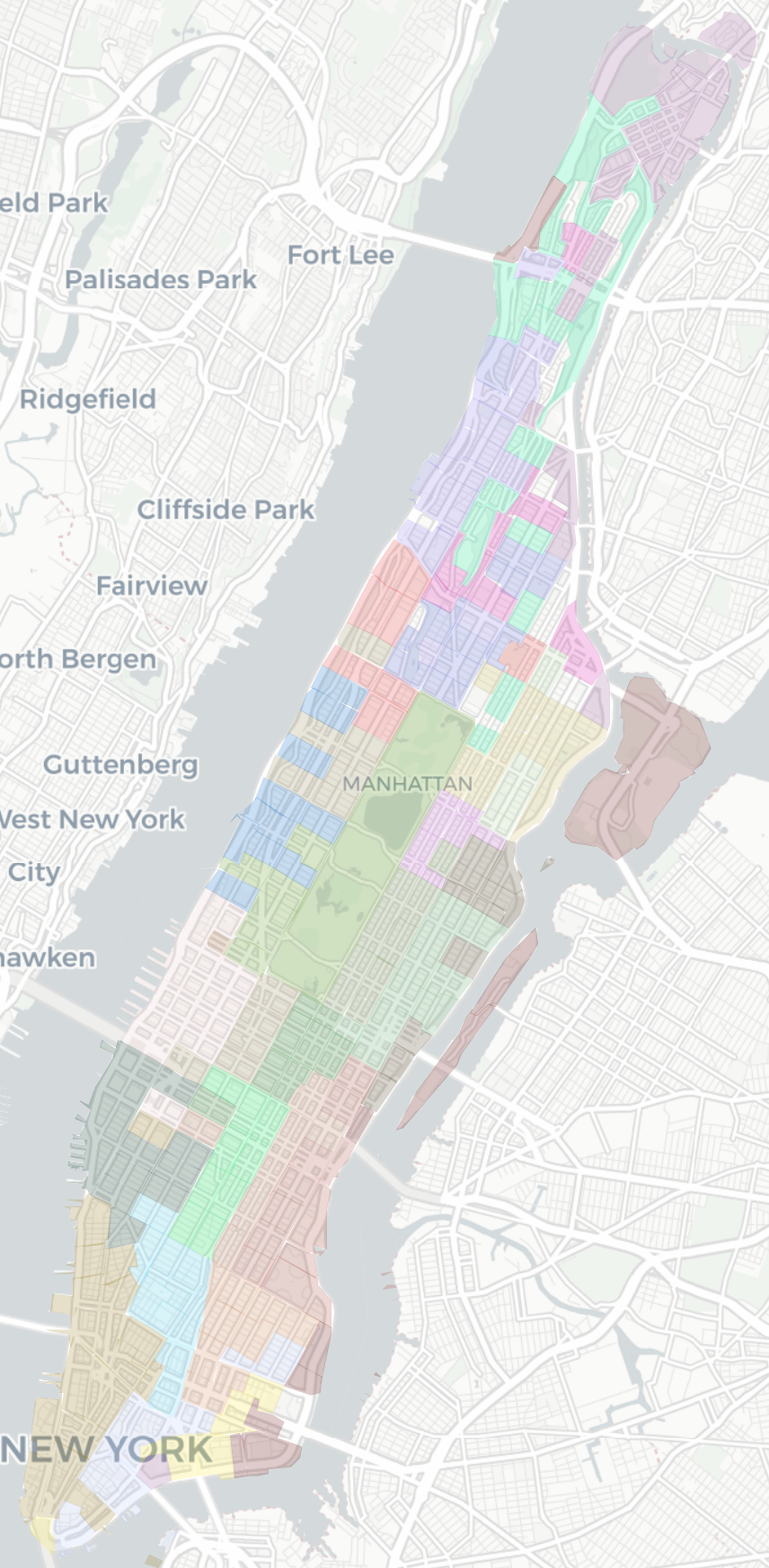}}
		\caption{\fontsize{6.6pt}{6.6pt}\selectfont{MVURE \cite{zhang2020multi}}}
		\label{fig:MVURE}
	\end{subfigure}
	\caption{Districts in Manhattan and region clusters.} 
	\label{fig:cluster}
	\vspace{-0.5cm}
\end{figure}

We first evaluate the learned embeddings in the land usage clustering task. As shown in the Fig. \ref{fig:cluster}, the borough of Manhattan is divided into 29 districts based on land usage. We conduct k-means clustering ($k$=29) with the region embeddings as inputs, and 
the regions in the same cluster are assigned with the same type.

We show the {\LCThree performance results} of different methods in Table~\ref{tab:CR}, where the best performance is in boldface and the second best is underlined. Overall, our method performs consistently better than all the baselines in terms of three metrics, and the margin reaches near 10\% for NMI and near 30\% 
for ARI and F-measure. Specifically, POI (TF-IDF) performs the worst among the baselines. A possible reason is that 
the high dimensional representation based on POI categories is inefficient for the clustering task. 
For HDGE and ZE-Mob,
the region attributes such as POIs are not considered, which may cause information loss. MVURE and M2GRL which consider multi-view region correlations {\LCThree perform} much better than single-view based methods. For MV-PN and CGAL, they achieve comparable performance because they adopt the same idea but different training strategies. 
In contrast, we use the contrastive learning to highlight the discriminate features, which help our model achieve the best {\LCThree performance}. 

We also visualize the clustering results of the methods 
that have the top ARI scores in Fig. \ref{fig:cluster}, where two regions in the same color are in the same cluster.
The cluster size is used as the index of colors here.
We can find that  the functional regions are weakly identified by M2GRL, and some clusters are mixed with each other especially in the middle right part. In contrast, ReMVC gives more satisfying identification with clearer cluster boundaries.
More visualization results can be found {\LCThree in a technical report \footnotemark[4]}.

\vspace{-0.25cm}
\subsection{Popularity Prediction}
Region popularity prediction is another commonly used task \cite{fu2019efficient}. Following this setting, we aggregate the check-in counts within each region as the ground truth of popularity, which can be found in \cite{yang2014modeling}. To evaluate the effectiveness of different region embedding methods, we use the Lasso regression model 
for this downstream task. All results are obtained by 5-Fold cross-validation. 

The results are also shown in Table \ref{tab:CR}. The proposed model performs consistently better than all the baselines in terms of three metrics. 
Among the baselines, MV-PN and CGAL achieve comparable results with the POI (TF-IDF) methods. A possible reason is that the mobility modeling module serves as an oversimplified regularization part in both methods, and the region representation is obtained by compressing POI features. These models cannot fully exploit multi-view correlations.
Because human mobility is highly related to region popularity, HDGE and ZE-Mob achieve good results on this task. {\LIANG{Compared with other multi-view learning methods MVURE and M2GRL, our method performs better with the inter-view contrastive learning module to propagate information across different views. 
}}


\vspace{-0.25cm}
\subsection{Module Evaluation} 
\label{ablation}
In this experiment, we further inspect how each module of  ReMVC affects the clustering and regression performance.

Specifically, by removing the POI encoder $G^p$ and Mobility encoder $G^m$ from ReMVC, we can obtain two simplified models w/o POI and w/o Mob, in which we train our model only based on one of the two views. We also disable the inter-view contrastive learning module and directly concatenate two embeddings without knowledge propagation in a simplified model w/o IV. Additionally, in order to verify the necessity of exploiting contrastive learning in the intra-view feature extraction, we replace it with an AutoEncoder using MSE loss and obtain the base model ReMVC-MSE. The encoder and decoder are implemented {\LCThree as} $G_{MLP}$ and the size of middle layer is {\LCThree set as} 32 for fair comparison. 


From the results {\LCThree shown} in Table~\ref{tab:ablation}, we have the following observations.
(1) ReMVC which considers multi-view information generally has better performance than the methods considering only single-view data, i.e., w/o POI and w/o Mob. The result demonstrates that each view has its own merit to enhance the region representation.
Meanwhile, the model w/o POI outperforms w/o Mob on both two tasks. {\LIANG{One possible reason is that POIs often suffer from data sparsity problem.}} 
(2) Our full model ReMVC brings around 20\% (in terms of ARI) and 17\% (in terms of MAE) improvement compared with the base model w/o IV, {\LIANG{showing that the proposed inter-view contrastive learning module can fully exploit the multi-view data and the importance of information propagation between views.}} (3) ReMVC outperforms ReMVC-MSE by a large margin especially in {\LCThree the} land usage clustering task. The results show that the feature reconstruction pretext task is not as effective as contrastive learning in region representation learning. Similar to the image data, a region can be identified by its distinct structures rather than its pixel-level details.   



In general, contrastive learning and multi-view cooperation are key components for our model. Due to the page limit, we present the experiment results of other ablation studies and parameters analysis {\LCThree in a technical report \footnotemark[4].}

\begin{table}[t!]
	\renewcommand\arraystretch{1.1}
	\centering
	\scriptsize
	\caption{Performance Comparison of Different Modules.}
	\vspace{-0.2cm}
	\resizebox{0.84\linewidth}{!}{
		\begin{tabular}{c|ccc|ccc}
			\toprule
			\multirow{2}{*}{Method} & \multicolumn{3}{c|}{{\it Land Usage Clustering}} & \multicolumn{3}{c}{{\it Popularity Prediction}}  \\
			\cmidrule{2-7}
			& NMI & ARI & F-measure & MAE & RMSE & $R^2$  \\ \midrule
			w/o POI & \underline{0.759} & \underline{0.430} & \underline{0.449} & 253.10 & 385.12 & 0.393  \\
			w/o Mob & 0.474 & 0.083 & 0.115 & 238.86 & 357.72 & 0.477  \\
			w/o IV & 0.723 & 0.393 & 0.405 & 229.77 & 343.26 & 0.518 \\
		    ReMVC-MSE & 0.517 & 0.127 & 0.151 & \underline{191.68} & \underline{299.91} & \underline{0.632} \\
			\cmidrule{1-7}
			ReMVC & \textbf{0.762} & \textbf{0.474} & \textbf{0.488} & \textbf{189.92} & \textbf{296.26} & \textbf{0.643} \\ \bottomrule
	\end{tabular}}
	\label{tab:ablation}
	\vspace{-0.4cm}
\end{table}

\vspace{-0.4cm}
\section{Conclusion}
In this paper, we study the problem of 
learning an embedding space for regions using multi-view information. 
We identify and introduce two guidelines for this problem: $i$) comparing a region with others within each view for effective representation extraction and $ii$) comparing a region with itself across different views for cross-view information sharing. 
{\LIANG{Based on the guidelines, we propose a novel multi-view joint learning model ReMVC for region embedding, which features in the schemes of 
intra-view and inter-view contrastive learning. 
}}
%
%
{\FinalOne Note that our 
contrastive learning structure can also be applicable to other domains. When this structure is applied to other domains, some implementation details (including the augmentation methods, loss functions, etc.) need to {\LCThree be} adjusted to better suit the target domain.}



\vspace{-0.2cm}
\bibliographystyle{IEEEtran}
\bibliography{related_works}





\end{document}